\documentclass[10pt,twocolumn,letterpaper]{article}

\usepackage{wacv}
\usepackage{times}
\usepackage{epsfig}
\usepackage{graphicx}
\usepackage{amsmath}
\usepackage{amssymb}
\usepackage{color}
\usepackage{multirow}
\usepackage{wrapfig,lipsum,booktabs}
\usepackage[algoruled,boxed,lined]{algorithm2e}
\SetAlFnt{\tiny}
\SetAlCapFnt{\scriptsize}
\SetAlCapNameFnt{\scriptsize}
\usepackage{adjustbox}
\usepackage{dirtytalk}

\def\wacvPaperID{85} 

\wacvfinalcopy 

\ifwacvfinal
\def\assignedStartPage{1} 
\fi

\ifwacvfinal
\usepackage[breaklinks=true,bookmarks=false]{hyperref}
\else
\usepackage[pagebackref=true,breaklinks=true,colorlinks,bookmarks=false]{hyperref}
\fi

\ifwacvfinal
\else
\fi

\begin{document}

\title{Assisting Scene Graph Generation with Self-Supervision}

\author{Sandeep Inuganti\\
{\tt\small sandeep.ipk@iith.ac.in}
\and
Vineeth N Balasubramanian\\
{\tt\small vineethnb@iith.ac.in}\\
\\
\hspace{-6cm}Department of Computer Science and Engineering\\
\hspace{-6cm}Indian Institute of Technology, Hyderabad, INDIA\\
}

\maketitle

\begin{abstract}
Research in scene graph generation has quickly gained traction in the past few years because of its potential to help in downstream tasks like visual question answering, image captioning, etc. Many interesting approaches have been proposed to tackle this problem. Most of these works have a pre-trained object detection model as a preliminary feature extractor. Therefore, getting object bounding box proposals from the object detection model is relatively cheaper. We take advantage of this ready availability of bounding box annotations produced by the pre-trained detector. We propose a set of three novel yet simple self-supervision tasks and train them as auxiliary multi-tasks to the main model. While comparing, we train the base-model from scratch with these self-supervision tasks, we achieve state-of-the-art results in all the metrics and recall settings. We also resolve some of the confusion between two types of relationships: geometric and possessive, by training the model with the proposed self-supervision losses. We use the benchmark dataset, Visual Genome to conduct our experiments and show our results. 

\end{abstract}

\section{Introduction}
The success of deep learning in performing basic tasks such as classification, detection, and segmentation has led to using deep neural networks in higher-level vision problems like visual scene understanding. Scene graph generation (SGG) being one of the most important problems in this category. A Scene graph not only gives us a visually grounded representation, but, it is also very useful in performing other higher-level tasks like evaluating~\cite{Anderson2016SPICESP} \& improving~\cite{Liu2016ImprovedIC} image captioning, visual question answering~\cite{Johnson2017InferringAE}, and image generation~\cite{Johnson2018ImageGF}. Therefore, an improvement in the performance of SGG will also help in the above mentioned higher-level tasks.

Given an image, a scene graph generation method constructs a directed graph, where the direction goes from the subject to object, and the edge between these two denotes the relationship they have. A scene graph is formally defined as: {\it Given an image $I$, let the set of object categories be $C$ and the set of relationship categories be $R$, now, a scene graph $G(I)$ is a tuple, $(O,E)$ where $O$ is a set of objects in the image $I$, with each $o \in C$, and $E \in O \times R \times O$ is a set of directed edges of the form $(s, r, o)$ where $s, o \in O$ and $r \in R$}. With an increase in interest in the SGG methods for visual scene understanding, many interesting methods were proposed in recent years. Most of these methods use pre-trained object detectors~\cite{Ren2015FasterRT} as their feature extractor in the first stage and then have further modules for graph generation and relationship classification. Some of these methods are described in the next section. 
In this work, our goal is to develop auxiliary tasks to assist scene graph generation in making better relationship detections. 

Manual annotation is expensive, especially in the age of deep learning. To deal with this problem, there has been increasing interest in recent years to learn representations from unlabeled data using self-supervision. Evident from some recent works like~\cite{Lee2019MultiTaskSO}, and ~\cite{Su2019BoostingSW}, that learning a supervised task with multi-task self-supervision losses boosts the performance of the base model on which these tasks are applied as auxiliary heads. These methods are also described in the next section. While these works improve upon the state-of-the-art (SOTA) methods for image classification and object detection, improving the SOTA in scene graph generation with self-supervision tasks has not been addressed yet. The challenges in introducing self-supervision for SGG are two-fold: 1) The graph-generation and relationship classification modules in the second stage of SGG require a different self-supervision than the existing tasks. 2) The SGG models are already computationally very intensive, so, the self-supervision tasks must not have heavy overhead over the main task. Therefore, to devise the self-supervision tasks, we take the advantage of the generic pipeline of a scene graph generation model. We use the bounding boxes produced by the object detector to compute labels for our self-supervision tasks. We propose three types of auxiliary self-supervision tasks: relative position classification, intersection over union prediction, and euclidean distance prediction. 
\begin{figure*}
\begin{center}
\includegraphics[width=57mm, height=22mm]{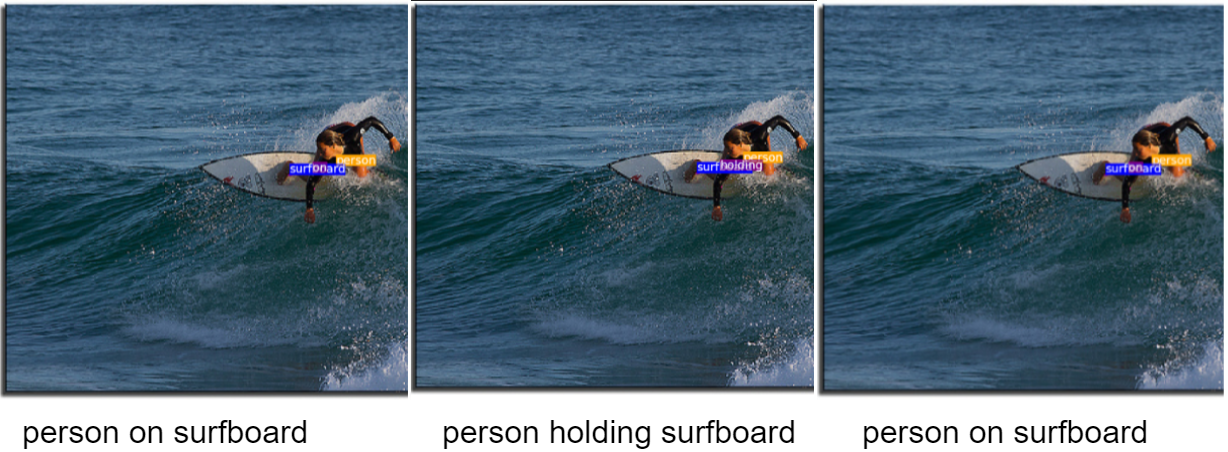}  \includegraphics[width=57mm, height=22mm]{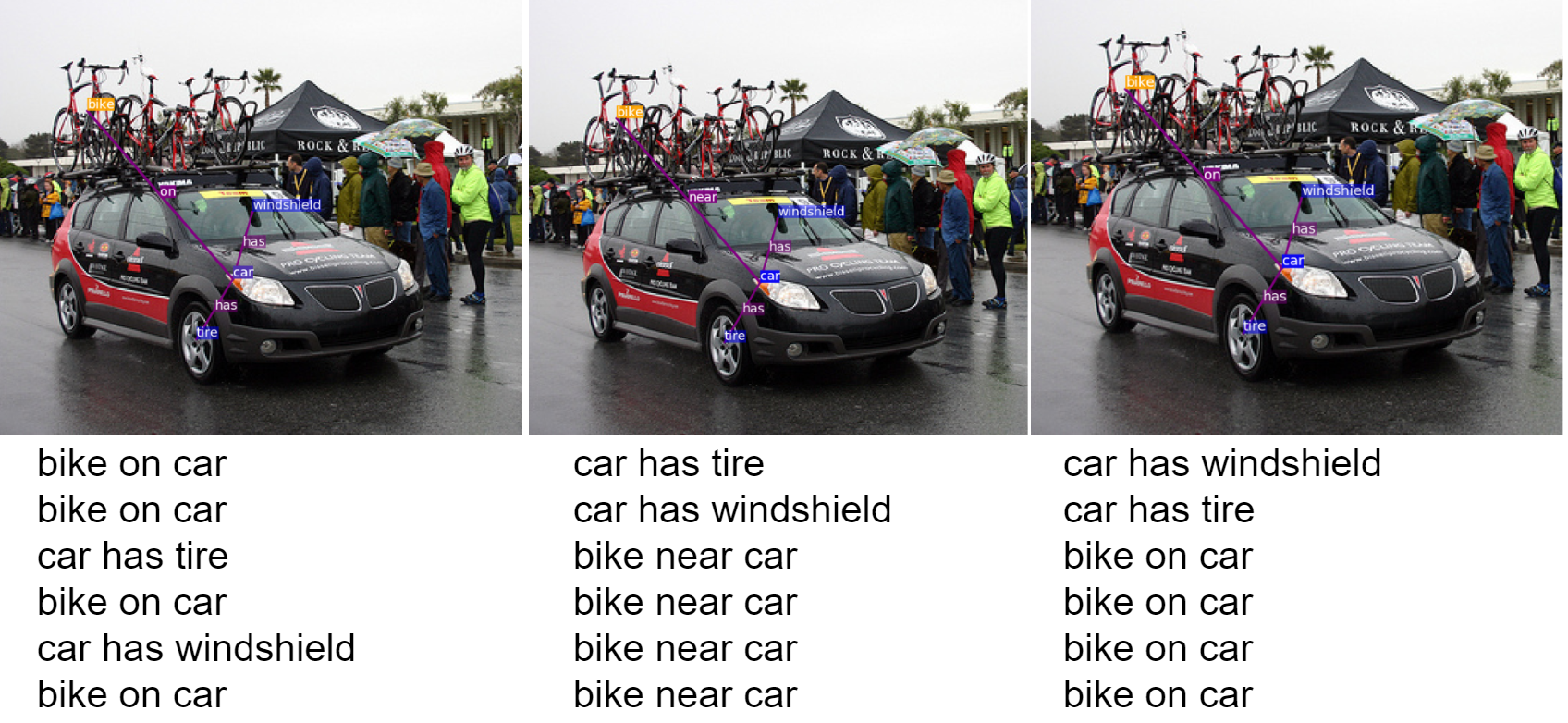}

\vspace{3mm}

\includegraphics[width=57mm, height=22mm]{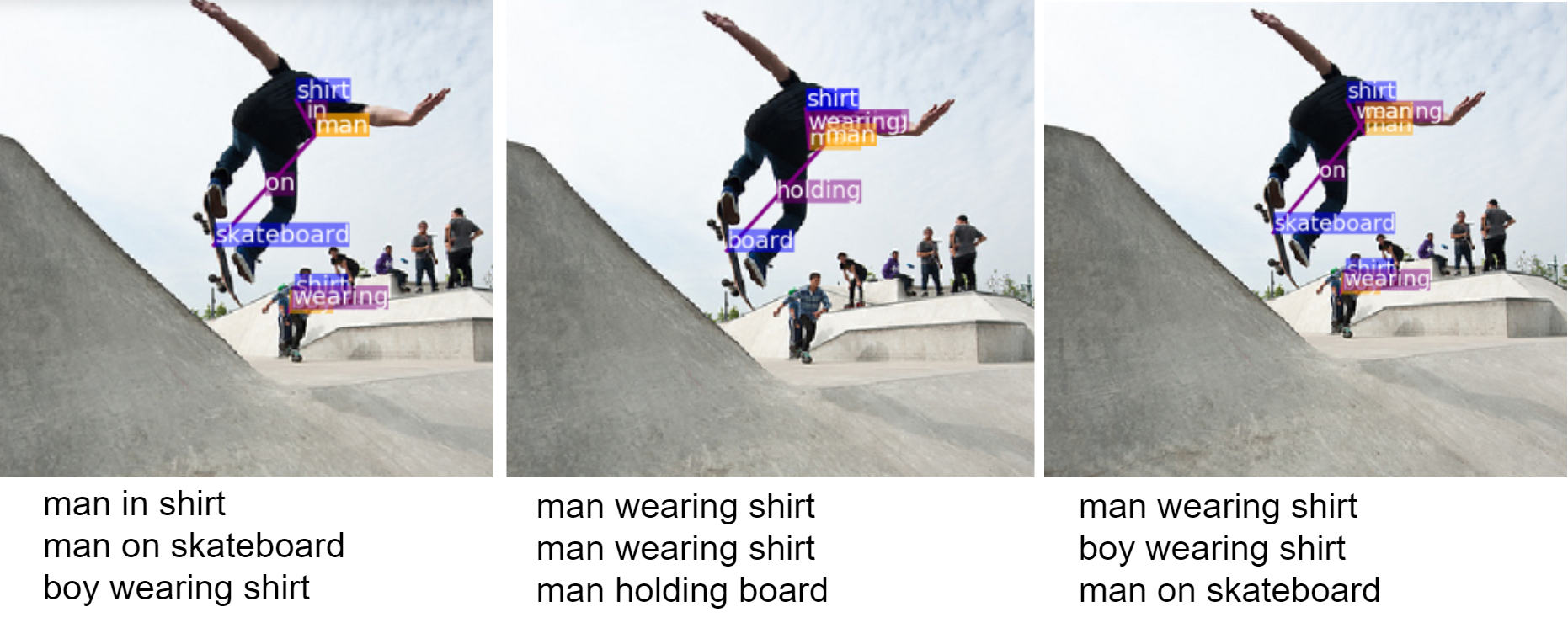}  \includegraphics[width=57mm, height=22mm]{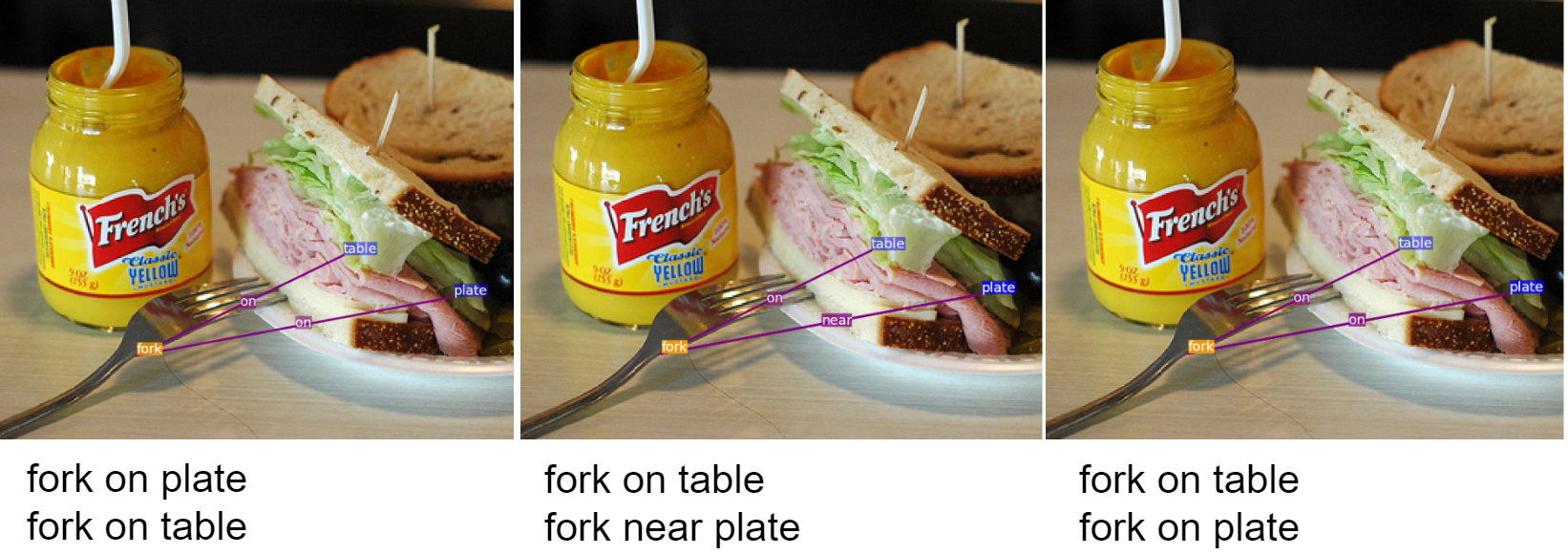}
\end{center}
\vspace{-0.3cm}
\caption{In each of the sub-figure, image on the left is ground truth, image in the middle is the predicted relationship by the base-model. With self-supervision, the predicted relationship is corrected in the image on the right. In the first sub-figure, the geometric relationship type (on) is confused with a possessive type (holding)}
\label{fig:confusion}
\vspace{-0.5cm}
\end{figure*}
\\
In addition to improving the performance of an SGG model, we also improve the graph generation qualitatively. An important observation on the scene graphs generated by the SOTA models~\cite{zhang2019vrd} is that they sometimes fail to discriminate between two relationship types: geometric (e.g under, on, etc.) and possessive (e.g holding, wearing etc). Often models confuse between these relationship types and predict incorrect relationships, as shown in figure~\ref{fig:confusion}. Our self-supervision tasks assist the models and help alleviate this problem. We provide some more example qualitative results in section~\ref{qualitativeres}, which show that self-supervision is able to resolve this confusion. The SOTA~\cite{zhang2019vrd} model also introduces three types of contrastive losses in their paper to improve the performance on the Visual Genome~\cite{Krishna2016VisualGC} dataset. But, we beat them with our simpler self-supervision losses.

Based on the results of the experiments conducted, we hypothesize that training the model on multi-task self-supervision losses will take advantage of joint feature learning, to learn features that have a better encoding of the spatial and visual information. Thereby, discriminating better between the geometric and possessive relationship types. 
To support our hypothesis, we analyze the t-SNE~\cite{Maaten2008VisualizingDU} visualizations of features learned by base-model and base-model with self-supervision in section~\ref{qualitativeres}. By various experiments and ablation studies, we show that our self-supervision tasks improve state-of-the-art performance. The contributions of this paper can be summarized as:
\\
1)To the best of our knowledge, this is the first attempt on providing self-supervision in the field of SGG. With almost no additional computational overhead, we aid the SGG task significantly. 
\\
2)We propose a set of three auxiliary self-supervision tasks to boost the performance of scene graph generation. These self-supervision tasks are a simple, elegant, and easy-to-implement way to improve the performance of SGG.
\\
3)We conduct various experiments to show the improvement in results, both qualitatively and quantitatively (+{\bf 5.4\%} recall@20 SGDET). We resolve the confusion between geometric and possessive relation types (+{\bf 2.5\%} recall@20 and +{\bf 5.8\%} recall@20 PREDCLS respectively). We also outperform the contrastive losses proposed in the SOTA model~\cite{zhang2019vrd} with our self-supervision losses.

We hope that our work can open a broader discussion around the use of self-supervision in more complex vision tasks including scene graph generation.
\section{Related Work}

{\bf Scene Graph Generation:}
Scene graph generation was introduced in the works by Ranjay Krishna et al.~\cite{Krishna2016VisualGC} and Johnson et al., 2015~\cite{Johnson2015ImageRU}. The utility of scene graphs in giving a structured representation, and helping in higher-level tasks has given rise to many scene graph generation methods proposed in recent years~\cite{Lu2016VisualRD,Xu2017SceneGG,Newell2017PixelsTG,Li2017ViPCNNAV,Li2017SceneGG,Krishna2016VisualGC,zellers2018scenegraphs,Yang2018GraphRF,zhang2019vrd,Dornadula2019VisualRA,Chen2019SceneGP}. Most of the works in scene graph generation have object detectors~\cite{Ren2015FasterRT} as their base, which give rise to the potential object proposals in the image. If there are $N$ object proposals then there are $N^2$ possible binary relationships. While methods like~\cite{zellers2018scenegraphs} attempt to classify all of those quadratic possibilities, most methods resort to random or heuristic sampling for choosing a subset out of these possible relationships. The existing scene graph generation methods can be broadly categorized into context-propagation and substructure-based methods. Context-based methods~\cite{Li2017ViPCNNAV,Xu2017SceneGG,Yang2018GraphRF} have techniques for context propagation through a candidate scene graph (generated from either sampling over $N^2$ possible relationships or using all of them) for refining the final relationship classification. Substructure-based methods~\cite{li2018fnet,zellers2018scenegraphs} analyze regularly appearing substructures in scene graphs. Johnson et al., 2018~\cite{Johnson2018ImageGF} also use scene graphs to generate images. They use GCNs ~\cite{Kipf2016SemiSupervisedCW} to parse the scene graphs into cascaded refinement networks (CRN)~\cite{Chen2017PhotographicIS} to generate images.
\cite{zellers2018scenegraphs} also categorize the types of relationships into three types: geometric, possessive, and semantic. 
\begin{table}
\begin{center}
\begin{adjustbox}{width=6.5cm,center}
\begin{tabular}{|c|c|c|c|}
\hline
Type & Examples & Classes & Instances\\ 
\hline
Geometric & above, behind, under & 15 & 228K\\
Possessive & has, part of, wearing & 8 & 186K\\
Semantic & carrying, eating, using & 24 & 39K\\
Others & for, from, made of & 3 & 2K\\ 
\hline
\end{tabular}
\end{adjustbox}
\end{center}
\caption{Relationship Types in Visual Genome dataset}
\label{table:reltypes}
\vspace{-0.5cm}
\end{table} 
Table~\ref{table:reltypes} shows the distribution of the relationship types in Visual Genome~\cite{Krishna2016VisualGC} dataset. 

We can see that geometric and possessive type relationships are the most common. Nearly half of the classes belong to semantic relationship type, but the instances of those relationships are significantly lesser. To deal with this problem, Zellers et al.~\cite{zellers2018scenegraphs} propose a frequency baseline which performs better than some existing scene graph generation methods, and it is entirely dependent on the distribution of objects and predicates in the dataset. Zhang et al.~\cite{zhang2019vrd} use a similar frequency baseline which forms a co-occurrence matrix of relationship class frequencies that occur between any two types of object classes. To deal with geometric and possessive types, they develop two modules: spatial and visual, further described in section~\ref{basemodel}.
While the SOTA methods~\cite{zhang2019vrd} perform well, they still have some confusion between geometric/possessive relation types. To solve this problem we introduce self-supervision for a better discerning capability. 
\\
\\
{\bf Self-Supervision:}
The data already contains some inherent structural information that can be utilized for feature learning. In recent years many self-supervised learning techniques were proposed to deal with image feature learning~\cite{Kingma2013AutoEncodingVB,Noroozi2016UnsupervisedLO,Radford2015UnsupervisedRL,Larsson2017ColorizationAA,Gidaris2018UnsupervisedRL}, and video feature learning~\cite{Vondrick2016GeneratingVW,Saito2016TemporalGA,Ahsan2018VideoJU,Fischer2015FlowNetLO}. The methods of self-supervision in images can be broadly classified into generative methods~\cite{Radford2015UnsupervisedRL} and contextual methods~\cite{Noroozi2016UnsupervisedLO,Gidaris2018UnsupervisedRL}, while self-supervision methods in videos can be broadly classified into generative methods~\cite{Vondrick2016GeneratingVW,Saito2016TemporalGA} and cross-modal methods~\cite{Fischer2015FlowNetLO,Arandjelovic2017LookLA}. A comprehensive survey of these self-supervision methods can be found in~\cite{Jing2019SelfsupervisedVF}.
Some of the recent works ~\cite{Su2019BoostingSW,Mangla2019ChartingTR,Lee2019MultiTaskSO} show that training supervision models with self-supervision losses gives a boost in performance. 

One work that could be considered closest to ours is~\cite{Gu2019SceneGG} by Gu et al, which regularizes scene graph generation using image reconstruction~\cite{Johnson2018ImageGF}. This work does not explicitly view this regularization as self-supervision, although one could implicitly view this so. However, the image reconstruction~\cite{Johnson2018ImageGF} method is a computationally expensive operation, and does not yield reliable results (which was supported by our own empirical studies of self-supervision).\\
\\
{\bf Multi-Task Learning:}
Along with self-supervision, recent efforts such as ~\cite{Su2019BoostingSW} and~\cite{Lee2019MultiTaskSO} also take advantage of viewing this from the perspectives of Multi-Task Learning (MLT). As the degree of multi-tasking increases~\cite{Baxter1997ABT}, MLT provides a good regularization between various tasks, thereby reducing the data required for a better generalization on the task. Multi-Task Learning has been influential in improving a number of vision tasks like object detection~\cite{Lee2019MultiTaskSO}, synthetic image generation~\cite{Ren2017CrossDomainSM}, depth estimation \& scene parsing~\cite{Xu2018PADNetMG}, etc. MLT methods can be broadly classified into two groups w.r.t the sharing of the parameters between different tasks: {1)\it Soft Parameter Sharing}: Here, each task has its own model with its own parameters. The methods in this category focus on how to design weight sharing with constraints (like a distance metric), to be imposed between the parameters. Some of the works in this category are DCNet~\cite{Trottier2018MultiTaskLB}, Cross-Stitch Network~\cite{Misra2016CrossStitchNF}, Sluice Networks~\cite{Ruder2017SluiceNL}, and partially shared multi-task CNN~\cite{Cao2018PartiallySM}; and {2)\it Hard Parameter Sharing}: In these methods, all the tasks share the same feature extractor then have a separate head to perform their own task. Therefore, appropriate tasks and loss functions should be used. Some of the works in this category are Mask R-CNN~\cite{He2017MaskR}, HyperFace~\cite{Ranjan2016HyperFaceAD}, and ResNetCrowd~\cite{Marsden2017ResnetCrowdAR}. Since our self-supervision tasks need to aid the main task (relationship detection) and thus improve the main backbone architecture, we choose the hard parameter sharing approach in this work. We now describe our methodology.
\begin{figure*}
\begin{center}
\includegraphics[width=120mm, height=50mm]{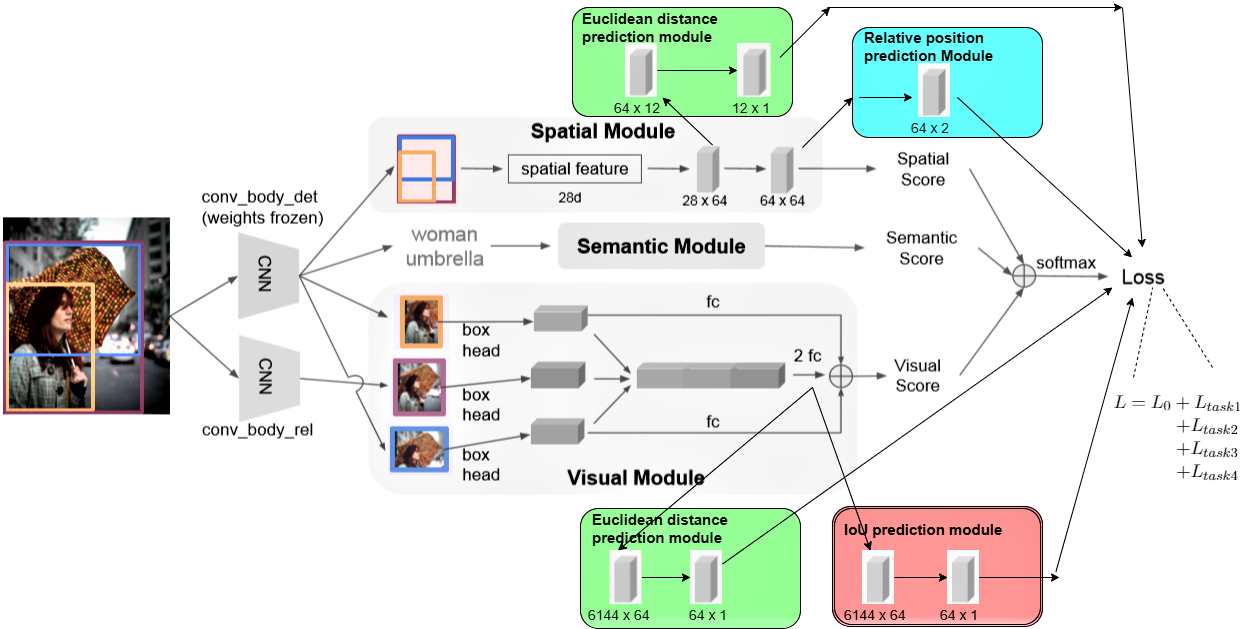}
\end{center}
   \caption{Our overall architecture. The SOTA base-model~\cite{zhang2019vrd} is in black-and-white, while the self-supervision tasks are added in color. Our self-supervision tasks attempt to assist the predictions of geometric and possessive relationship types.}
\label{fig:fullmodel}
\end{figure*}
\begin{table*}
\begin{center}
\label{table:selfsup}
\begin{adjustbox}{width=12cm,center}
\begin{tabular}{|c|c|c|c|}
\hline
Task & Description & Type & Loss\\ 
\hline
\hline
\#1 & Relative Position classification & Geometric & Multi-Label Binary Cross Entropy\\
\#2, \#4 & Euclidean distance regression & Geometric/Possessive & Mean Squared Error \\
\#3 & Intersection over Union regression & Possessive & Mean Squared Error\\
\hline
\end{tabular}
\end{adjustbox}
\end{center}
\caption{The proposed Self-supervision tasks}
\vspace{-0.4cm}
\end{table*}
\section{Methodology}
Before we present our self-supervision tasks, we begin with the preliminaries from the SOTA~\cite{zhang2019vrd} SGG model which deals with the geometric and possessive relationship types. Without loss of generality, our contributions herein would be applicable to any other SGG method that uses similar relationship types.

\vspace{-6pt}
\subsection{Preliminaries}
\vspace{-6pt}
The {\bf spatial module} computes spatial features based on relative positions of subjects and objects from the bounding box proposals. 
The equations for calculating these feature vectors are as follows: 
\begin{equation}
\Delta(b_1,b_2) = \Big{<} \frac{x_1 - x_2}{w_2}, \frac{y_1 - y_2}{h_2},  \log\frac{w_1}{w_2},   \log\frac{h_1}{h_2} \Big{>}
\label{deltabox}
\end{equation}
\begin{equation}
c(b) = \Big{<} \frac{x}{W}, \frac{y}{H},  \frac{x + w}{W},   \frac{y + h}{H}, \frac{wh}{WH} \Big{>}
\label{normbox}
\end{equation}
where $b_1$ and $b_2$ are the two object bounding boxes in the form the tuple $\big{<}x, y, w, h\big{>}$, and W and H are the dimensions of the image. Finally the spatial feature is constructed as follows:
\begin{equation}
\big{<} \Delta(b_s, b_{pred}), \Delta(b_{pred}, b_o),  \Delta(b_s, b_o),  c(b_s),  c(b_o) \big{>}
\label{sptfeat}
\end{equation}
where $b_s$ and $b_o$ are the subject and object bounding box respectively, and $b_{pred}$ is the tightest bounding box of the subject and object under consideration. These features are passed through a multi-layer perceptron (MLP) to get geometric relationship scores. The subject and object ROI features are extracted by an object CNN applied on the pooled ROI feature maps produced by the object detection pipeline. Simultaneously, relationship ROI features are extracted from a parallel relation CNN. These features are passed through the {\bf visual module} to get possessive relationship scores (see fig~\ref{fig:fullmodel}). The final prediction vector is computed by adding all three module's scores and applying softmax on top of the combined score:
\begin{equation}
    p = softmax(f_{spt} + f_{vis} + f_{sem})
\end{equation}
where $f_{spt}$, $f_{vis}$, and $f_{sem}$ are the spatial, visual, and semantic module's scores respectively and $p$ is the final relationship prediction vector for the current subject-relationship-object triplet.
\label{basemodel}
\subsection{Self-supervision Tasks}
\label{sstaks}
\vspace{-6pt}
Self-supervision tasks are generally formulated on the dataset by creating artificial labels for tasks like jigsaw~\cite{Noroozi2016UnsupervisedLO}, rotation prediction~\cite{Gidaris2018UnsupervisedRL}, etc. These labels can be generated automatically without any manual annotations. 
In the case of scene graph generation, the base of most of the methods is object detection. We leverage this fact of pre-trained object detectors which output proposals for subjects and objects. We don't use any additional information, and simply use these generated proposals to compute labels for all of our self-supervision tasks. 
We use four simple self-supervision losses to deal with two types of relationship types: possessive and geometric types. Since the relationship detection model from Zhang et al. provides two separate modules: visual and spatial to deal with the before mentioned relationship types respectively, we choose to provide the self-supervision tasks to these two modules. We describe the two types of self-supervision used below:\vspace{-1pt}
\\
{\bf Geometric self-supervision: }\\
\noindent \textbf{\textit{Task 1: Relative position classification}:} Relative position classification refers to a multi-label binary classification problem with labels as left/right and up/down position of the object bounding box w.r.t subject bounding box. The motivation for this task comes from observing the examples of geometric type relationships like \say{above}, \say{below}, etc. These relationship classes can be inferred using relative spatial information, and we can get this information by comparing the centroids of the bounding boxes.  

\noindent \textbf{\textit{Task 2: Euclidean distance prediction}:} This task is formulated as a regression task where the spatial module should predict the distance between the centroids of the subject and object bounding boxes. We assume that the prediction of Euclidean distance is beneficial in learning representations for the nearness of an object, which could aid in resolving the geometric/possessive confusion. Therefore, we use the same task for both types of self-supervision.\\
\\
{\bf Possessive self-supervision: }\\
\noindent \textbf{\textit{Task 3: Intersection over Union (IoU) prediction}:} IoU prediction refers to a regression task with labels as computed IoUs between the subject and object bounding boxes under consideration. The motivation for possessive self-supervision also comes from a similar observation as in the geometric case, in relationships like \say{has}, \say{part of}, etc. Intersection over union prediction serves as a basic task for knowing how much overlap is present between the two bounding boxes and further helps in determining possessivity. Also, as described above, we use Euclidean distance prediction as one more task for possessive self-supervision as {\bf Task 4}. 


\subsection{Auxiliary Heads}
\vspace{-6pt}
The head for Relative Position Classification comes from the features of final hidden layer activations of the spatial module and comes out as a two-layer MLP with a two-dimensional sigmoid activated output for multi-label classification. Therefore, we use a multi-label binary cross-entropy (BCE) loss for this task: 
\\
\begin{equation}
\begin{split}
       L_{task1} = \ell(x, y) = - 1/n\cdot\sum_{i=1}^{n}\sum_{c=1,2}[ y_{ic} \cdot \log x_{ic} \\
       + (1 - y_{ic}) \cdot \log (1 - x_{ic})]
\end{split}
\end{equation}
where $x$ and $y$ are the output of the auxiliary head and label respectively, $c$ is the class number, and there are two classes: left/right and top/bottom, and $n$ is the no. of relationship proposals considered in the given image. 

Both {Euclidean distance prediction} tasks are formulated as regression tasks. Both heads from spatial and visual modules come out from the pre-final hidden layer activations as a three-layer MLP with a positive scalar output. Therefore, we use mean squared error loss function for both tasks \#2 and \#4 as described in the following equation:
\begin{equation}
        L_{task2} = \ell(x, y) = 1/n\cdot\sum_{i=1}^{n}( x_i - y_i )^2
\label{mseloss}
\end{equation}
The {IoU prediction} task is also formulated as a regression task. The head for this task comes out of the visual module from the pre-final hidden layer as a three-layer MLP producing a positive scalar output, we use mean squared error loss as described in equation~\ref{mseloss}. We also formulated this task with a BCE loss, but the results were similar ($\pm 0.01$ SGDET) to this regression formulation. The total loss function for the output of the relationship detector is described in following equation~\ref{totalloss}:
\begin{equation}
        L = L_0 + L_{task1} + L_{task2} + L_{task3} + L_{task4}
\label{totalloss}
\end{equation}
where $L$ is the total loss, $L_0$ is the cross-entropy loss for relationship classification, $L_{task}$s are the self-supervision losses as described in Table~\ref{table:selfsup},
and the model is described in Figure~\ref{fig:fullmodel}. We evaluate the proposed methodology on the benchmark dataset~\cite{Krishna2016VisualGC} for SGG, as well as run ablation studies on the above tasks and losses in Sections~\ref{quantres} to~\ref{ablations}.

\section{Experiments and Results}
\label{expandresults}
In this section, we study the performance of the proposed self-supervision tasks and compare them to state-of-the-art methods. 
We conduct ablation studies on the proposed self-supervision tasks with various combinations to study how each task is contributing to the performance. We also conduct a model-specific ablation study to see if the intended type of self-supervision is performing as expected. Lastly, we study the usefulness of the proposed tasks on another top-performing SGG model, FactorizableNet~\cite{li2018fnet}, to study their relevance to other kinds of SGG methods.

\subsection{Implementation Details}
\label{imp_details}
We evaluate our contributions on the Visual Genome (VG) dataset as given in~\cite{zellers2018scenegraphs}. The VG dataset has 62,723 training images and 26,446 testing images. We note here that the Visual Genome dataset is an extension of the Visual Relationship Detection (VRD)~\cite{Lu2016VisualRD} dataset. While some earlier efforts (especially older work) in SGG have used the VRD dataset, more recent work such as Graph R-CNN~\cite{Yang2018GraphRF}, and MotifNet~\cite{zellers2018scenegraphs} have focused solely on Visual Genome since it is an extension of the VRD dataset. We hence also use the newer \& larger VG, instead of VRD, for our studies. The object CNN (conv\_body\_det as shown in Figure~\ref{fig:fullmodel}) weights are frozen, but the relationship CNN (conv\_body\_rel) weights are made trainable, and they are initialized to the object detector's convolutional layer weights. 

Given the maximum value in the classification probability score vectors: $p_{sub}$, $p_{obj}$, and $p_{rel}$ of subject, object and relationship respectively, we multiply all of these three ($p_{sub} * p_{obj} * p_{rel}$) and rank the $\big<sub, rel, obj\big>$ detections. We then compute the recall$@$K using the Scene Graph Detection (SGDET), Scene Graph Classification (SGCLS), and Predicate Classification (PREDCLS) as the evaluation metrics with the same definitions as given in~\cite{Xu2017SceneGG}. 
We use Stochastic Gradient Descent (SGD) as our optimizer, with an initial learning rate of 0.005. All the implementations are in Pytorch 1.0, and while training, we use 4 Nvidia Tesla P100 GPUs (batch size = No. of GPUs used). For the training of base-model with self-supervision losses, on 62,723 images, it takes $\sim$ 18 hours; which is lesser than the base model with contrastive losses ~\cite{zhang2019vrd} which takes around 20-21 hours. During training, we sample 512 pairs out of the $512 * 512$ possible relationships, and during testing, we use 100 object proposals and consider all of the $100 * 100$ possible relationships. In comparison, the baseline model ~\cite{zhang2019vrd} uses 8 GPUs and samples 2048 pairs out of the $512 * 512$ possible relationships during training. This is a significantly higher setting than ours but, we still outperform them with our lower experiment setting. We also implement our self-supervision tasks on another model, FactorizableNet~\cite{li2018fnet}. Here, we train the model on 15,000 training images, and at test time, we use 5,000 images to report the results. This model starts with an initial learning rate of 0.005, and SGD is used as the optimizer. These experiments are further described in section~\ref{ablations}.

\subsection{Quantitative Results}
\label{quantres}
\begin{table*}[h]
\begin{center}
\begin{adjustbox}{width=12cm,center}
\begin{tabular}{|c|c|c|c|c|c|c|c|c|c|}
\hline
\multirow{2}{*}{Model Name}   & \multicolumn{3}{c|}{SGDET} & \multicolumn{3}{c|}{SGCLS} & \multicolumn{3}{c|}{PREDCLS} \\ \cline{2-10} 
                              & R@20    & R@50    & R@100  & R@20    & R@50    & R@100  & R@20     & R@50    & R@100   \\ \hline
Frequency Baseline~\cite{zellers2018scenegraphs}           & 17.7    & 23.5    & 27.6   & 27.7    & 32.4    & 34.0   & 49.4     & 59.9    & 64.1    \\ 
Frequency Baseline + Overlap~\cite{zellers2018scenegraphs}  & 20.1    & 26.2    & 30.1   & 29.3    & 32.3    & 32.9   & 53.6     & 60.6    & 62.2    \\ 
VRD~\cite{Lu2016VisualRD}                           & -       & 0.3     & 0.5    & -       & 11.8    & 14.1   & -        & 27.9    & 35.0    \\ 
Associative Embedding~\cite{Newell2017PixelsTG}         & 6.5     & 8.1     & 8.2    & 18.2    & 21.8    & 22.6   & 47.9     & 54.1    & 55.4    \\ 
IMP~\cite{Xu2017SceneGG}               & -       & 3.4     & 4.2    & -       & 21.7    & 24.4   & -        & 44.8    & 53.0    \\ 
IMP + Frequency Baseline & 14.6    & 20.7    & 24.5   & 31.7    & 34.6    & 35.4   & 52.7     & 59.3    & 61.3    \\ 
MotifNet - NoContext~\cite{zellers2018scenegraphs}          & 21.0    & 26.2    & 29.0   & 31.9    & 34.8    & 35.5   & 57.0     & 63.7    & 65.6    \\ 
MotifNet - LeftRight~\cite{zellers2018scenegraphs}          & 21.4    & 27.2    & 30.3   & 32.9    & 35.8    & 36.5   & 58.5     & 65.2    & 67.1    \\ \hline
Base Model$^*$~\cite{zhang2019vrd}                    & 20.61   & 27.85   & 32.38  & 36.07   & 36.73   & 36.74  & 66.94    & 68.50   & 68.52   \\ 
Base Model + Constrastive Losses$^*$~\cite{zhang2019vrd}                    & 20.85   & 27.88   & 32.31  & 36.02   & 36.66   & 36.67  & 66.77    & 68.28   & 68.29   \\ 
Base Model + Self-supervision$^*$(ours) & {\bf21.73}   & {\bf28.28}   & {\bf{}32.56}  & {\bf36.36}   & {\bf37.01}   & {\bf37.03}  & {\bf67.15}   & {\bf68.85}   & {\bf68.87}   \\ \hline
Base Model$^\dagger$~\cite{zhang2019vrd}                    & 20.8   & 28.1   & 32.5  & 36.1   & 36.7   & 36.7  & 66.7    & 68.3   & 68.3   \\ 
Base Model + Constrastive Losses$^\dagger$~\cite{zhang2019vrd}                    & 21.1   & 28.3   & 32.7  & 36.1   & 36.8   & 36.8  & 66.9    & 68.4   & 68.4  \\ \hline 
\end{tabular}
\end{adjustbox}
\end{center}
\caption{SGG results on the VG dataset. $*$ shows that the models were trained from scratch. $\dagger$ shows the reported results from~\cite{zhang2019vrd}, in which the compute settings were significantly higher than ours. Please refer to Section~\ref{imp_details} for details of implementation.}
\label{table:comparison}
\end{table*}
We compare the base model trained with our self-supervision losses, with the previous methods in scene graph generation. We train the base model ~\cite{zhang2019vrd} and base model + self-supervision from scratch with settings as given in Section~\ref{imp_details}. The results are shown in Table~\ref{table:comparison}. We beat existing state-of-the-art models in all metrics and all recall settings. Moreover, we significantly outperform in SGDET recall@20 metric. It is the most important because it is the hardest metric that judges both the model's detection and classification capabilities. Recall@20 is important because it gives us the correct predictions within the top 20 predictions (while R@50 and R@100 allow upto top 50 and 100 predictions).

The Frequency Baseline model~\cite{zellers2018scenegraphs} is similar to the semantic module described in Section~\ref{basemodel}. As we can see it outperforms some of the existing methods by large. Frequency Baseline + Overlap means that, while the frequency counts are being pre-computed by the Frequency method, we increment the frequency counts only if the subject and object proposals have an overlap. We can also observe that adding Frequency Baseline to Iterative Message Passing (IMP)~\cite{Xu2017SceneGG} improves the method drastically, but still performs lower than our method. MotifNet~\cite{zellers2018scenegraphs} is the model proposed by Zellers et al., who introduced this Frequency Baseline. From the results, we can infer that self-supervision significantly assists SGG. It can also be inferred that using different modules for different relationship types is beneficial. Base Model + Self-supervision refers to using all the self-supervision tasks (tasks \#1, \#2, \#3, and \#4) to do multi-task learning on the Base Model. ~\cite{zhang2019vrd} also propose 3 types of graphical contrastive losses, and infer that these losses have very minor gains on the base model for the VG dataset. We conducted experiments with these contrastive losses and results were similar with or without them. We hence report our results on the model without these losses, without loss in generality, when training the base model with our self-supervision tasks.

\subsection{Qualitative Results}
\label{qualitativeres}
\begin{figure*}
\vspace{-0.3cm}
\begin{center}
\includegraphics[width=57mm, height=22mm]{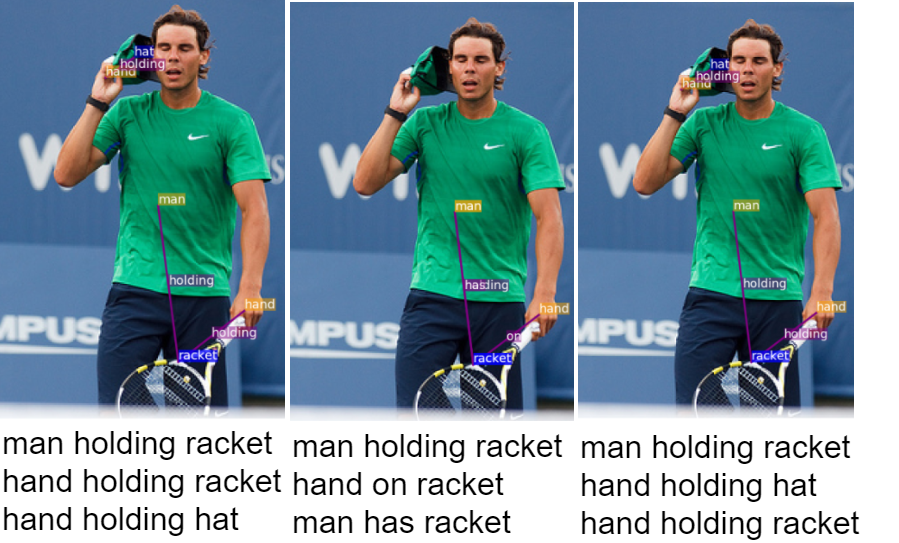}\includegraphics[width=57mm, height=22mm]{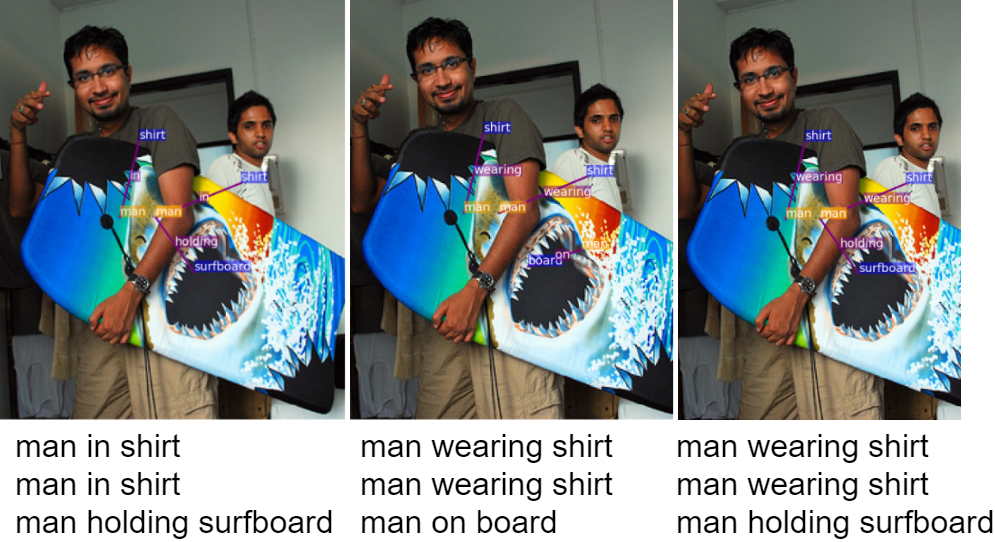}

\vspace{3mm}

\includegraphics[width=57mm, height=22mm]{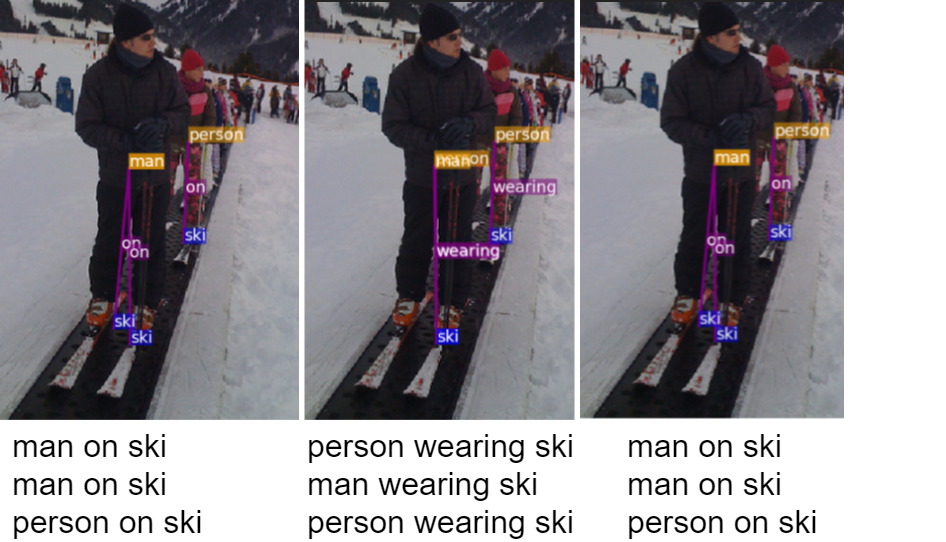}\includegraphics[width=57mm, height=22mm]{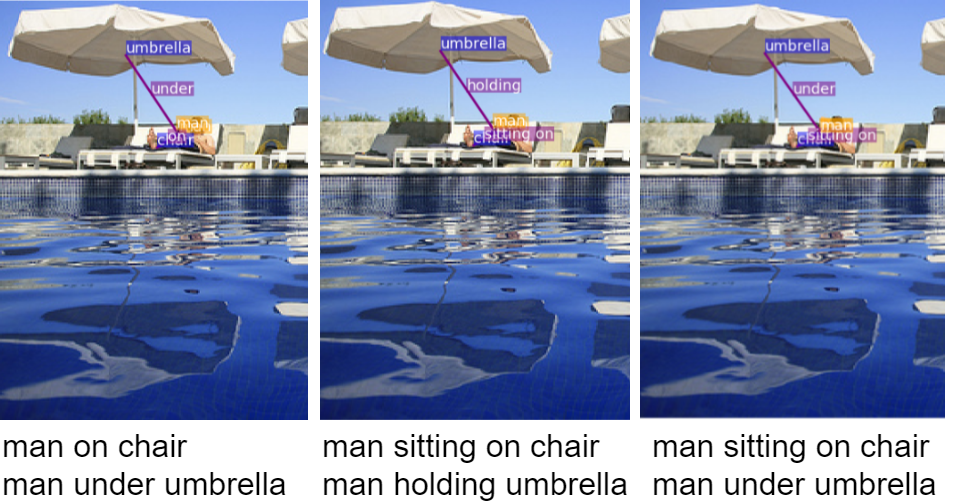}
\end{center}
\vspace{-0.2cm}
   \caption{Each of the four sub-figures shows the output given by the ground truth, base-model, and base-model with self-supervision, respectively. The first row shows the resolution of geometric to possessive relationship type achieved by our method, the second shows the vice-versa.}
\label{fig:qualitative_geometric}
\vspace{-0.3cm}
\end{figure*}

\begin{figure}
\begin{center}
\includegraphics[width=58mm, height=26mm]{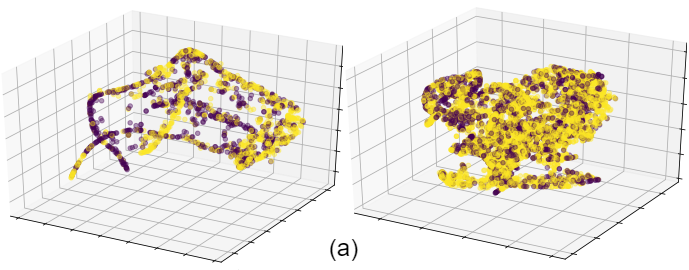}
\vspace{2mm}
\includegraphics[width=58mm, height=26mm]{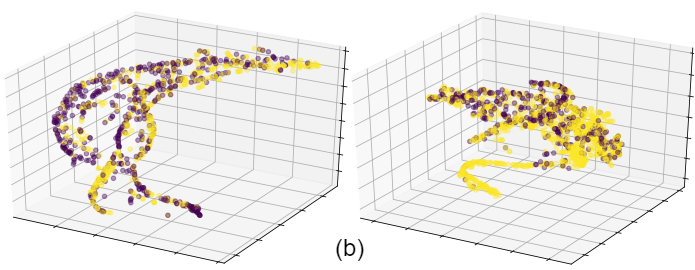}
\end{center}
\vspace{-0.3cm}
  \caption{(a) is plotted at a threshold of 0.5 with 100 random examples each for base model with and without self-supervision respectively, (b) is plotted at a threshold of 0.8 with 1000 random examples. Yellow represents possessive relation type, purple represents geometric type \textit{(best viewed in colour)}}
\label{fig:tsne_vis}
\end{figure}

Sample qualitative results are provided in Figure~\ref{fig:qualitative_geometric}'s top and bottom row for the possessive and geometric types of relationships respectively. We observe that the model with self-supervision resolves the confusion between the possessive and geometric relationship types. In the top row of Figure~\ref{fig:qualitative_geometric}, we see that the base model is confused and predicts a geometric relation: \say{on} between \say{racket} and \say{hand}, with self-supervision it is resolved to a possessive relation: \say{holding}. Similarly, in the bottom row, the base model predicts a possessive relation: \say{holding} as the relation between \say{man} and \say{umbrella}, with self-supervision it is resolved to a geometric relation: \say{under}. Moreover, we also observe that at a threshold of 0.5 the base-model predicts {\bf 5.3} times more non-trivial relationships than with self-supervision. This shows that the base-model is confused and predicts relatively more relationships at an indecisive threshold of around 0.5. But, if we increase the threshold to 0.9, the trend flips, now the base-model with self-supervision predicts {\bf 3.4} times more non-trivial relationships, which shows that with self-supervision the base-model detects relatively higher confident predictions ({\it in\_image} relationship is trivial).
\vspace{-0.3cm}
\paragraph{t-SNE Visualizations: } To analyze how the addition of self-supervision performs in discriminating the geometric from possessive relationships, we plot the t-SNE visualizations of the final layer relationship features from the base-model with and without self-supervision respectively in Figures~\ref{fig:tsne_vis}(a) and ~\ref{fig:tsne_vis}(b). In Figure~\ref{fig:tsne_vis}(a), the two plots are plotted by randomly sampling 100 images at a threshold confidence score of 0.5. The base model gives around 6700 geometric and possessive relation detections, and with self-supervision, the model gives around 1700 detections. We can see that the base model with self-supervision discriminates better between geometric and possessive relations, while the base model seems to be confused. Now, the same can be seen in Figure~\ref{fig:tsne_vis}(b), where we increase the confidence score threshold to 0.8 for 1000 randomly sampled images. The base model with self-supervision gives around 1100 geometric and possessive relation detections, and without gives around 1400 detections, reiterating our earlier claim of higher confidence detections with the proposed approach.

\begin{figure}
\begin{center}
\includegraphics[width=60mm, height=32mm]{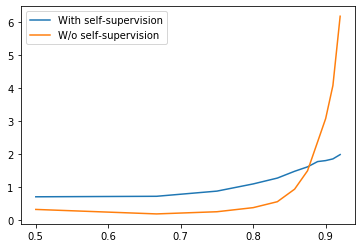}
\end{center}
\vspace{-0.3cm}
  \caption{Variation of $\alpha$ with threshold, over the whole test set (26446 images). x-axis: threshold, y-axis: $\alpha$}
\label{fig:stats}
\end{figure}

\begin{table}
\begin{center}
\begin{adjustbox}{width=9cm,center}
\begin{tabular}{|c|c|c|c|c|c|c|c|c|c|c|c|}
\hline
Threshold &  0.5 & 0.667 & 0.75 & 0.80 & 0.833 & 0.857 & 0.875 & 0.889 & 0.9 & 0.91 & 0.92\\
\hline
$\alpha_{s}$ & 0.718 & 0.733 & 0.890 & 1.108 & 1.285 & 1.491 & 1.625 & 1.785 & 1.815 & 1.866 & 1.997\\

$\alpha_{ns}$ & 0.335 & 0.199 & 0.265 & 0.389 & 0.569 & 0.949 &  1.507 & 2.383 & 3.088 & 4.090 & 6.189\\
\hline
\end{tabular}
\end{adjustbox}
\end{center}
\caption{$\alpha_{s}$ denotes the geometric to possessive ratio with self-supervision and $\alpha_{ns}$ is without self-supervision}
\label{table:statis}
\vspace{-0.5cm}
\end{table}
Another interesting observation in our results is that the base model predicts more detections of the possessive type. However, with the assistance of our simple self-supervision, the model maintains a better ratio as in the case of 0.5 threshold. This trend is depicted in Table~\ref{table:statis}, where we discuss how the ratio (denoted by $\alpha$) of the predicted geometric to possessive relations changes as we
increase the threshold on the whole test set (26446 images). In Figure~\ref{fig:stats} we depict this ratio in a graph. We can see that the $\alpha$ for the base-model with self-supervision is between 1.0 to 2.0 (which is in line with the dataset statistics in Table \ref{table:reltypes}). However, without self-supervision, the ratio shows a big variation.




\subsection{Ablation Studies}
\label{ablations}
We conduct ablations on the proposed self-supervision tasks by studying different combinations of the self-supervision tasks, and understanding how they contribute. We study the following combinations in these studies (Task \#1 and \#3 are also included as separate tasks in the results):\\
{- Combination1}: 
task \#1 + task \#3\\
{- Combination2}: 
task \#1 + task \#2 + task \#3\\
{- Combination3}: 
task \#1 + task \#3 + task \#4\\
{- Combination4}: 
task \#1 + task \#2 + task \#3 + task \#4

Combination1 is chosen to see how effectively these two main self-supervision tasks help in improving the baseline. After that, Euclidean distance prediction (task \#2, \#4) is added to each of the geometric and possessive self-supervision tasks to see if this task helps in improving any performance. The results of these studies are given in Table~\ref{table:ablationtasks}. Table~\ref{table:ablationtasks} shows that Combination1 gains good amount of recall in SGDET, SGCLS, and PREDCLS over the base model. But when we compare Combination1 with Combinations2 and 3, which adds Euclidean distance as one additional task, the performance gains are ambiguous. This points to the inference that the gains with Euclidean distance prediction over Combination1 is not convincing. Finally, when we use all the self-supervision tasks (in combination 4), we get better results than other combinations in all metrics.
\begin{table*}
\begin{center}
\begin{adjustbox}{width=12cm,center}
\begin{tabular}{|c|l|l|l|l|l|l|l|l|l|}
\hline
\multicolumn{1}{|l|}{\multirow{2}{*}{Combination}} & \multicolumn{3}{c|}{SGDET} & \multicolumn{3}{c|}{SGCLS} & \multicolumn{3}{c|}{PREDCLS} \\ \cline{2-10} 
\multicolumn{1}{|l|}{}                             & R@20    & R@50    & R@100  & R@20    & R@50    & R@100  & R@20     & R@50    & R@100   \\ \hline
only task\#1                                                & 21.65   & 28.12   & 32.50  & 36.31   & 36.98   & 37.00  & 67.11    & 68.77   & 68.79   \\ 
only task\#3                                                & 21.72   & 28.18   & 32.51  & 36.30   & 36.96   & 36.98  & 67.12    & 68.78   & 68.79   \\ 
1                                                & 21.69   & 28.21   & 32.49  & 36.29   & 36.96   & 36.97  & 67.10    & 68.77   & 68.79   \\ 
2                                                & 21.70   & 28.21   & 32.51  & 36.29   & 36.95   & 36.97  & 67.02    & 68.72   & 68.73   \\ 
3                                                & 21.61   & 28.24   & 32.53  & 36.27   & 36.94   & 36.95  & 67.02    & 68.71   & 68.72   \\ 
4                                                & {\bf21.73}   & {\bf28.28}   & {\bf{}32.56}  & {\bf36.36}   & {\bf37.01}   & {\bf37.03}  & {\bf67.15}   & {\bf68.85}   & {\bf68.87}   \\ \hline
\end{tabular}
\end{adjustbox}
\end{center}
\caption{Ablation studies on the proposed self-supervision tasks}
\label{table:ablationtasks}
\end{table*}

\begin{table*}
\begin{center}
\begin{adjustbox}{width=12cm,center}
\begin{tabular}{|c|c|c|c|c|c|c|c|c|c|}
\hline
\multirow{2}{*}{Module Used}          & \multicolumn{3}{c|}{SGDET} & \multicolumn{3}{c|}{SGCLS} & \multicolumn{3}{c|}{PREDCLS} \\ \cline{2-10} 
                                      & R@20    & R@50    & R@100  & R@20    & R@50   & R@100   & R@20    & R@50    & R@100    \\ \hline
Semantic + Visual                    & 20.09       & 26.99       & 31.46      & 35.59   & 36.25  & 36.26        & 66.12       & 67.66       & 67.67        \\ 
Semantic + Visual + Spatial          & 20.61   & 27.85   & 32.38  & 36.07   & 36.73   & 36.74  & 66.94    & 68.50   & 68.52        \\ 
Semantic + Visual + Task \#3 + \#4 & 21.32   & 27.92   & 32.27  & 36.18       & 36.86      & 36.87       & 66.82       & 68.51       & 68.53        \\ 
\begin{tabular}[c]{@{}c@{}}Semantic + Visual + Spatial\\ Task \#1 + \#2 + \#3 + \#4\end{tabular}  & {\bf21.73}   & {\bf28.28}   & {\bf{}32.56}  & {\bf36.36}   & {\bf37.01}   & {\bf37.03}  & {\bf67.15}   & {\bf68.85}   & {\bf68.87} \\ \hline
\end{tabular}
\end{adjustbox}
\end{center}
\caption{Model specific ablation studies on the proposed self-supervision tasks}
\label{table:ablationmodules}
\end{table*}

\begin{table}
\begin{center}
\begin{adjustbox}{width=9cm,center}
\begin{tabular}{|c|l|l|l|l|l|l|}
\hline
\multirow{2}{*}{Model}          & \multicolumn{3}{c|}{Possessive types} & \multicolumn{3}{c|}{Geometric types} \\ \cline{2-7} 
& R@20    & R@50    & R@100  & R@20    & R@50    & R@100   \\ \hline
Base Model  & 40.67   & 40.96   & 40.96   & 73.70   & 74.60   & 74.61   \\ 
Base Model + Self-supervision    & {\bf43.04}   & {\bf43.39}   & {\bf43.40}  & {\bf75.49}   & {\bf76.55}   & {\bf76.57}   \\ \hline
\end{tabular}
\end{adjustbox}
\end{center}
\caption{Results on PREDCLS metric for possessive and geometric types}
\label{table:relationablation}
\end{table}

\begin{table}
\begin{center}
\begin{adjustbox}{width=9cm,center}
\begin{tabular}{|c|l|l|l|l|l|l|l|l|l|}
\hline
{\multirow{2}{*}{Combination}}          & \multicolumn{3}{c|}{SGDET} & \multicolumn{3}{c|}{SGCLS} & \multicolumn{3}{c|}{PREDCLS} \\ \cline{2-10} 
\multicolumn{1}{|l|}{}  & R@20    & R@50    & R@100  & R@20    & R@50    & R@100  & R@20     & R@50    & R@100   \\ \hline
FNet~\cite{li2018fnet}  & 5.06   & 7.54   & 9.46   & 8.44   & 13.02   & 16.81    & 30.82    & 45.65   & 57.66   \\ 
FNet + task\#1 + \#3    & 5.09   & {\bf7.68}   & 9.60  & 8.60   & 13.17   & 16.82  & 30.51    & 45.74   & 57.37   \\ 
FNet + task\#1 + \#3 +\#4    & {\bf5.14}   & 7.64   & {\bf{}9.73}  & {\bf8.86}   & {\bf13.34}   & {\bf17.21}  & {\bf30.90}   & {\bf45.84}   & {\bf57.68}   \\ \hline
\end{tabular}
\end{adjustbox}
\end{center}
\caption{Results on FactorizableNet. In this case, Tasks \#2 and \#4 turn out to be the same because, this model doesn't have two separate modules to give these two tasks separately, and we hence use one of these tasks.}
\vspace{-0.5cm}
\label{table:ablationfnet}
\end{table}
For a deeper understanding, we also performed ablation studies on the base model, and report the results in Table~\ref{table:ablationmodules}. Here, we check to see if our self-supervision tasks are providing any improvement when we use individual modules of the base model and compare them with their versions with additional self-supervision losses. We can see that our self-supervision tasks add significant improvement to the visual module in the base model.\footnote{During backpropagation, the gradients from visual module are also used in updating the \say{conv\_body\_rel} weights (in Figure~\ref{fig:fullmodel}) by default. It is hence not possible to perform an ablation study just for the spatial module.} We gain significantly in all the metrics and all the recall settings. This corroborates our claim that multi-task self-supervision is aiding the individual modules to perform better. We also experimented on weighing the self-supervision losses, but it did not show any significant improvement over just using Equation~\ref{totalloss}. We also provide an ablation on the relationship prediction metric by relationship type (possessive vs geometric) in Table~\ref{table:relationablation}. These results show that the model with self-supervision also gives better results quantitatively.
\vspace{-0.3cm}
\paragraph{Experiments on FactorizableNet: } In order to study the generalizability of our approach, we also studied how the proposed self-supervision losses can be integrated with other SGG methods - FactorizableNet (FNet)~\cite{li2018fnet}, in particular, which is a fast scene graph generation model with strong results proposed by Li et al. This method has a better computational efficiency for graph generation by proposing a subgraph-based connection graph for representing the scene graph during inference. First, a bottom-up clustering method is used to factorize the entire candidate-graph into subgraphs (each subgraph contains many objects and a subset of their relationships). Subsequently, by replacing these relationship representations with a lesser number of subgraph features, the computation is significantly reduced. We encourage the interested reader to refer~\cite{li2018fnet} for further details. Here, we add the self-supervision tasks after the Predicate-Inference (SRI) module in their model. Table~\ref{table:ablationfnet} shows the ablation studies. We can see a similar trend as in the case of ablations on the base model.

\vspace{-0.5cm}
\section{Conclusion and Future work}
\vspace{-0.3cm}
In this work, we proposed a set of simple self-supervision tasks for the first time in scene graph generation. We take inspiration from the classification of types of relationships proposed by Zellers et al.~\cite{zellers2018scenegraphs} and design self-supervision tasks to improve the performance of the state-of-the-art~\cite{zhang2019vrd}. We evaluate our tasks on the Visual Genome dataset and conduct various experiments and ablation studies. We outperform the current scene graph generation models using multi-task learning with these self-supervision tasks. We also provide qualitative examples where self-supervision resolved confusion between relationship types. One future research direction could be testing the base model trained with self-supervision in a few-shot setting for less frequent relationship classes, as well as developing more advanced self-supervision tasks that combine relationship types.



{\small
\bibliographystyle{ieee_fullname}
\bibliography{egbib}
}

\end{document}